# Recursive Least Squares Policy Control with Echo State Network

Chunyuan Zhang*
School of Computer Science and Technology
Hainan University
Haikou, China
email: zcy7566@126.com

Chao Liu
School of Computer Science and Technology
Hainan University
Haikou, China
email: lcdyx0618@126.com

Qi Song
School of Computer Science and Technology
Hainan University
Haikou, China
email: songqihnu@163.com

Jie Zhao
School of Computer Science and Technology
Hainan University
Haikou, China
email: zhaojie@lonelyme.com

*Abstract*—The echo state network (ESN) is a special type of recurrent neural networks for processing the time-series dataset. However, limited by the strong correlation among sequential samples of the agent, ESN-based policy control algorithms are difficult to use the recursive least squares (RLS) algorithm to update the ESN's parameters. To solve this problem, we propose two novel policy control algorithms, ESNRLS-Q and ESNRLS-Sarsa. Firstly, to reduce the correlation of training samples, we use the leaky integrator ESN and the mini-batch learning mode. Secondly, to make RLS suitable for training ESN in mini-batch mode, we present a new mean-approximation method for updating the RLS correlation matrix. Thirdly, to prevent ESN from over-fitting, we use the $L_1$ regularization technique. Lastly, to prevent the target state-action value from overestimation, we employ the Mellowmax method. Simulation results show that our algorithms have good convergence performance.

*Keywords—reinforcement learning, echo state network, recursive least squares, policy control algorithm*

## I. INTRODUCTION

Reinforcement learning (RL) is an important type of machine learning methods. By interacting with the unknown environment, the agent tries to learn the optimal control polices at different states for maximizing the cumulative expected return [1]. In the last three decades, to deal with the sequential decision problems with large discrete or continuous state space, RL algorithms mainly use the linear approximation technique to calculate the state-action values. Whereas, this technique has limited approximation ability and requires users to extract features manually. In recent years, it is gradually replaced by with the popularity of deep neural networks (DNNs) in RL, and the deep RL (DRL) [2,3] has become the research hotspot of RL. However, although DNNs are powerful, most of them are difficult to be trained. The echo state network (ESN) [4,5] is a special type of recurrent neural networks (RNNs). It has many merits of RNNs such as the good approximation ability and the automatic feature extracting ability. What is more, it can be trained efficiently by the least squares (LS) or the recursive least squares (RLS), since it only needs to train the parameters from the input layer to the output layer and from the reservoir layer to the output layer. Therefore, we will focus on how to use ESN to design the policy control algorithms in this paper.

In the past two decades, many researchers have done some work on policy control algorithms by using ESN to approximate the state-action values of the agent. Szita et al. [6] propose an ESN-Sarsa algorithm for the partially observable Markov decision process (POMDP) task, and prove the convergence of our algorithm. Chatzidimitriou and Mitkas [7] present an augmented topological neural evolution methods to train ESN, and take it as the approximator of the state-action values. Engedy and Horváth [8] use ESN and the incremental Gaussian mixture network to propose a new RL system model, which can solve the optimal control problem in the partially observable continuous state or action space. Whitley et al. [9] propose an improved ESN trained by a novel neuron selection method, and use it to solve RL tasks. Shi et al. [10] present an ESN-Q learning algorithm to solve the battery control problem. Chang and Futagami [11] propose a new algorithm by using the convolutional ESN to extract the features and using the evolutionary method to select the action of the agent, which has good performance demonstrated by some RL tasks. However, the training efficiency and learning performance of these algorithms need to be further improved, since their ESNs don't use the mini-batch training mode and don't use the LS or RLS training method.

In RL, the least squares policy iteration (LSPI) algorithm is probably the most efficient algorithm, which is first proposed by Lagoudakis et al. [12,13]. By making full use of the historical samples, it employs the LS method to solve the weight parameters directly. On this basis, Xu et al. [14] present a novel kernel-based LSPI algorithm by using the kernel sparsification technique. Zhou et al. [15] propose a batch LSPI algorithm, and prove its convergence theoretically. This algorithm stores the samples generated by agent online, and uses them in batches to update the weight parameters. Compared with the traditional policy control algorithms such as Q-learning and Sarsa, the above LSPI algorithms have faster convergence speed and better learning performance. Whereas, they requires to compute the inverse of the autocorrelation matrix at each iteration. In addition, they use the offline or semi-offline learning mode, which makes them unsuitable for the online RL control tasks. Buşoniu et al. [16] propose a so-called "online" LSPI algorithm, but it is still a semi-offline learning algorithm in essence. Until now, we have not found a true online policy control algorithm based on LS or RLS.

*Corresponding author. This paper is supported by the Natural Science Foundation of China (61762032, 11961018).

In response to the above problems, we propose two ESN-based RLS policy control algorithms, namely, ESNRLS-Q and ESNRLS-Sarsa. Our main contributions are as follows: 1) We employ the mini-batch learning mode and the leaky integrator ESN (LiESN) [17], which reduce the correlation of the training samples and make RLS able to be used in the ESN-based policy control algorithms. 2) We present a new mean-approximation method to update the RLS autocorrelation matrix, which can reduce the computational complexity and make the RLS algorithm suitable for training ESN in the mini-batch mode. 3) We present an $L_1$ regularization method to prevent ESN from over-fitting. 4) We employ the Mellowmax method [18] to prevent the target state-action values from overestimation. 5) We demonstrate the effectiveness of our algorithms by using both MDP and POMDP tasks.

## II. BACKGROUND

### A. Deep Policy Control Algorithms

RL is mainly used to solve sequential decision-making problems, which are usually formulated as MDPs [1]. An MDP is a five-tuple $<\mathcal{S}, \mathcal{A}, P, r, \gamma>$, where $\mathcal{S}$ and $\mathcal{A}$ denote the state space and the action space, $P(s'|s,a) \in [0,1]$ and $r(s'|s,a) \in \mathcal{R}$ denote the state-transition probability and the immediate reward from the state $s$ to the subsequent state $s'$ by taking the action $a$, and $\gamma \in (0,1]$ is the discount factor. The goal of RL is to learn the optimal policy $\pi^*$ for maximizing the cumulative expected return $J(\pi)$, that is, $\pi^* = \text{argmax}_\pi J(\pi)$. Whereas, $J(\pi)$ is hardly calculated directly, since $P(s'|s,a)$ and $r(s'|s,a)$ are unknown in RL. For a given state $s$, $a$ is determined by the control policy $\pi$, and $s'$ and $r$ can only be obtained by the agent's interaction with the environment. In practical applications, the policy control algorithms generally use the state-action value $Q^\pi(s,a)$ to measure the performance of $\pi$ when initial state and action are $s$ and $a$. In this paper, we assume that $\mathcal{S}$ is continuous and $\mathcal{A}$ is discrete. For this kind of MDP problems, $Q^\pi(s,a)$ is often approximated by DNNs in recent years, and a lot of novel policy control algorithms such as DQN [2] and the deep Sarsa [19] are proposed.

The DQN algorithm can be reviewed as follows. First, at the current time step $t$, the agent uses the $\epsilon$-greedy policy to select a random action $a_t$ with probability $\epsilon$ or otherwise set $a_t = \text{argmax}_{a \in \mathcal{A}} Q^\pi(s_t, a)$, where $Q^\pi(s_t, a)$ is approximated by the policy network as

$$Q^\pi(s_t, a) \approx Q(s_t, a; \Theta_{t-1}) \qquad (1)$$

where $s_t$ and $\Theta_{t-1}$ are the input and the parameter of the policy network. Then, after taking action $a_t$, the agent moves to the state $s'_t$, gets reward $r_t$, and stores $(s_t, a_t, s'_t, r_t)$ in the experience replay buffer $\mathcal{D}$. Next, using the mini batch $\mathcal{M}_t = \{(\hat{s}_{t,i}, \hat{a}_{t,i}, \hat{s}'_{t,i}, \hat{r}_{t,i})\}_{i=1,\cdots,M}$ sampled from $\mathcal{D}$, the loss of the policy network can be calculated as

$$L(\Theta_{t-1}) = \frac{1}{2M} \left\| Q^\pi(\widehat{S}_t, \widehat{a}_t) - Q(\widehat{S}_t, \widehat{a}_t; \Theta_{t-1}) \right\|_F^2 \qquad (2)$$

where $\widehat{S}_t = [\hat{s}_{t,1}, \cdots, \hat{s}_{t,M}]^T$, $\widehat{a}_t = [\hat{a}_{t,1}, \cdots, \hat{a}_{t,M}]^T$, and $Q^\pi(\widehat{S}_t, \widehat{a}_t)$ is the target value, which is approximated by the target network as

$$Q^\pi(\widehat{S}_t, \widehat{a}_t) = \hat{r}_t + \gamma \max_{a \in \mathcal{A}} Q(\widehat{S}'_t, a; \widetilde{\Theta}) \qquad (3)$$

where $\hat{r}_t = [\hat{r}_{t,1}, \cdots, \hat{r}_{t,M}]^T$, $\widehat{S}'_t = [\hat{s}'_{t,1}, \cdots, \hat{s}'_{t,M}]^T$, and $\widetilde{\Theta}$ is the parameter of the target neural network, which is copied from the policy network every some time steps or episodes. Note that $Q(\hat{s}'_{t,i}, a; \widetilde{\Theta}) = 0$ if $\hat{s}'_{t,i}$ is an absorbing state. Last, $\Theta_{t-1}$ can be updated by some optimization algorithms such as SGD and Adam. If using SGD, the update rule is

$$\Theta_t = \Theta_{t-1} - \alpha \nabla_{\Theta_{t-1}} L(\Theta_{t-1}) \qquad (4)$$

The deep Sarsa algorithm is similar to the DQN algorithm. Distinct from DQN, deep Sarsa is an on-policy algorithm, and its $Q^\pi(\widehat{S}_t, \widehat{a}_t)$ is approximated as

$$Q^\pi(\widehat{S}_t, \widehat{a}_t) = \hat{r}_t + \gamma Q(\widehat{S}'_t, a_{t+1}; \widetilde{\Theta}) \qquad (5)$$

where $a_{t+1}$ is the action of the agent at $s_{t+1}$. Note that $Q(\hat{s}'_{t,i}, a_{t+1}; \widetilde{\Theta}) = 0$ if $\hat{s}'_{t,i}$ is an absorbing state.

### B. Leaky Integrator ESN

ESN is a special RNN, which is generally made up of an input layer, a reservoir layer and an output layer. Among all of the ESN variants, the leaky integrator ESN (LiESN) is probably most widely used. Suppose the current input series $\mathcal{I}_t = \{x_t^{(k)}\}_{k=1,\cdots,T}$, the output $h_t^{(k)}$ of the reservoir layer in a LiESN is calculated as

$$h_t^{(k)} = (1-\eta) h_t^{(k-1)} + f((W^i)^T x_t^{(k)} + \eta (W^r)^T h_t^{(k-1)} + b^r) \qquad (6)$$

where $f(\cdot)$ is the activation function, $W^i$ is the parameter matrix from the input layer to this layer, $W^r$ is the parameter matrix inside this layer, $b^r$ is the bias, and $\eta \in (0,1]$ is the leaky rate. The output $y_t^{(k)}$ of the output layer in a LiESN is calculated as

$$y_t^{(k)} = (W_{t-1}^o)^T \left[ (x_t^{(k)})^T, (h_t^{(k)})^T \right]^T + b_{t-1}^o \qquad (7)$$

where $W_{t-1}^o$ is the parameter matrix from the input layer and the reservoir layer to this layer, and $b_{t-1}^o$ is the bias. Different from general RNNs, $W^i$ and $b^r$ are determined by the random initialization, $W^r$ is often determined by the spectral radius method [4], and only $W_{t-1}^o$ and $b_{t-1}^o$ are required to be trained. In order to facilitate subsequent discussion and simplify the expression, we define $\Theta_{t-1} = [(W_{t-1}^o)^T, b_{t-1}^o]^T$ and $u_t^{(k)} = \left[ (x_t^{(k)})^T, (h_t^{(k)})^T, 1 \right]^T$, and rewrite (8) as

$$y_t^{(k)} = \Theta_{t-1}^T u_t^{(k)} \qquad (8)$$

$\Theta_{t-1}^T$ is generally solved or updated by LS or RLS.

### III. ESNRLS-Q AND ESNRLS-SARSA

#### A. Algorithm Derivation

In this subsection, we will introduce the derivation of ESNRLS-Q and ESNRLS-Sarsa. Since the latter is similar to the former, we will mainly introduce the former, and then introduce the latter briefly.

Similar to DQN, ESNRLS-Q also uses the experience replay, the mini-batch learning mode and the $\epsilon$-greedy policy. Different from DQN, ESNRLS-Q uses two LiESNs as the policy network and the target network, and uses RLS to update the policy network. In addition, each sample in the experience replay buffer $\mathcal{D}$ of ESNRLS-Q is a series, which is made up of $T$ successive state-action transitions. If the number of the successive transitions is less than $T$ when the agent arrives in the absorbing state, ESNRLS-Q will use the last state-action transition to fill the sample.

Firstly, we introduce the forward learning of ESNRLS-Q. Suppose the current mini batch $\mathcal{M}_t = \{(\hat{s}_{t,i}^{(k)}, \hat{a}_{t,i}^{(k)}, \hat{s}'^{(k)}_{t,i}, \hat{r}_{t,i}^{(k)})\}_{i=1,\cdots,M;k=1,\cdots,T}$ sampled from $\mathcal{D}$. By using (6), the output $\mathbf{H}_t^{(k)}$ of the reservoir layer in the policy network can be calculated as

$$\mathbf{H}_t^{(k)} = (1-\eta)\mathbf{H}_t^{(k-1)} + f(\hat{\mathbf{S}}_t^{(k)}\mathbf{W}^i + \eta\mathbf{H}_t^{(k-1)}\mathbf{W}^r + \mathbf{1} \times (b^r)^\top) \quad (9)$$

where $\hat{\mathbf{S}}_t^{(k)} = [\hat{s}_{t,1}^{(k)}, \cdots, \hat{s}_{t,M}^{(k)}]^\top$, and $\mathbf{1} \in \mathcal{R}^M$ is a full-one vector. Then, by using (7) and (8), the output $Q(\hat{\mathbf{S}}_t^{(k)}, \mathcal{A}; \Theta_{t-1})$ of the policy network can be calculated as

$$Q(\hat{\mathbf{S}}_t^{(k)}, \mathcal{A}; \Theta_{t-1}) = \mathbf{U}_t^{(k)}\Theta_{t-1} \quad (10)$$

where $\mathbf{U}_t^{(k)} = [\hat{\mathbf{S}}_t^{(k)}, \mathbf{H}_t^{(k)}, \mathbf{1}]$. On this basis, we can easily get $Q(\hat{\mathbf{S}}_t^{(k)}, \hat{a}_t^{(k)}; \Theta_{t-1})$, where $\hat{a}_t^{(k)} = [\hat{a}_{t,1}^{(k)}, \cdots, \hat{a}_{t,M}^{(k)}]^\top$. Similarly, $Q(\hat{\mathbf{S}}_t^{(k)}, \mathcal{A}; \widetilde{\Theta})$ can be calculated by the targeted network. As a result, $Q^\pi(\hat{\mathbf{S}}_t^{(k)}, \hat{a}_t^{(k)})$ can be calculated as

$$Q^\pi(\hat{\mathbf{S}}_t^{(k)}, \hat{a}_t^{(k)}) = \hat{r}_t^{(k)} + \gamma \max_{a \in \mathcal{A}} Q(\hat{\mathbf{S}}_t'^{(k)}, a; \widetilde{\Theta}) \quad (11)$$

where $\hat{r}_t^{(k)} = [\hat{r}_{t,1}^{(k)}, \cdots, \hat{r}_{t,M}^{(k)}]^\top$. Note that $Q(\hat{s}_{t,i}'^{(k)}, a; \widetilde{\Theta}) = 0$ if $\hat{s}_{t,i}'^{(k)}$ is an absorbing state.

Secondly, we introduce the backward learning of ENRLS-Q, that is, how to use RLS to update $\Theta_{t-1}$. For simplifying expression, we define $Q^\pi(\hat{s}_{n,i}^{(k)}, a) = 0$ and $Q(\hat{s}_{n,i}^{(k)}, a; \Theta)$ if $a \neq \hat{a}_{n,k}^{(k)}$. Then, define the LS weighted loss function as

$$\hat{L}(\Theta) = \frac{1}{2MT}\sum_{n=1}^{t}\sum_{k=1}^{T}\lambda^{t-n}\|\boldsymbol{\delta}_n^{(k)}\|_F^2 \quad (12)$$

where $\boldsymbol{\delta}_n^{(k)} = Q^\pi(\hat{\mathbf{S}}_n^{(k)}, \mathcal{A}) - Q(\hat{\mathbf{S}}_n^{(k)}, \mathcal{A}; \Theta)$, and $\lambda \in (0,1]$ is the forgetting factor. Then, the parameter learning of the policy network can be summarized as

$$\Theta_t = \operatorname{argmin}_\Theta \hat{L}(\Theta) \quad (13)$$

Let $\nabla_\Theta = \partial \hat{L}(\Theta)/\partial \Theta$. By the chain rule, $\nabla_\Theta$ is

$$\nabla_\Theta = -\frac{1}{MT}\sum_{n=1}^{t}\sum_{k=1}^{T}\lambda^{t-n}(\mathbf{U}_n^{(k)})^\top \boldsymbol{\delta}_n^{(k)} \quad (14)$$

Let $\nabla_\Theta = \mathbf{0}$. Then, we can obtain

$$\Theta_t = \mathbf{A}_t^{-1}\mathbf{B}_t \quad (15)$$

where

$$\mathbf{A}_t = \frac{1}{MT}\sum_{n=1}^{t}\sum_{k=1}^{T}\lambda^{t-n}(\mathbf{U}_n^{(k)})^\top \mathbf{U}_n^{(k)} \quad (16)$$

$$\mathbf{B}_t = \frac{1}{MT}\sum_{n=1}^{t}\sum_{k=1}^{T}\lambda^{t-n}(\mathbf{U}_n^{(k)})^\top Q^\pi(\hat{\mathbf{S}}_t^{(k)}, \mathcal{A}) \quad (17)$$

Rewrite the above two equations into the following recursive forms

$$\mathbf{A}_t = \lambda\mathbf{A}_{t-1} + \frac{1}{MT}\sum_{k=1}^{T}(\mathbf{U}_t^{(k)})^\top \mathbf{U}_t^{(k)} \quad (18)$$

$$\mathbf{B}_t = \lambda\mathbf{B}_{t-1} + \frac{1}{MT}\sum_{k=1}^{T}(\mathbf{U}_t^{(k)})^\top Q^\pi(\hat{\mathbf{S}}_t^{(k)}, \mathcal{A}) \quad (19)$$

However, we cannot obtain $\mathbf{A}_t^{-1}$ by the Sherman-Morrison formula [20] directly, since the last term in the right hand side of (18) is a sum of matrix product.

Next, we propose a mean-approximation method to solve this problem. Considering that all states in a series are similar and the last terms in the right hand side of (18) and (19) are the mean of the product sum, we rewrite (18) and (19) as follows

$$\mathbf{A}_t \approx \lambda\mathbf{A}_{t-1} + \bar{\boldsymbol{u}}_t\bar{\boldsymbol{u}}_t^\top \quad (20)$$

$$\mathbf{B}_t \approx \lambda\mathbf{B}_{t-1} + \bar{\boldsymbol{u}}_t(\bar{Q}_t^\pi)^\top \quad (21)$$

where $\bar{\boldsymbol{u}}_t$ and $\bar{Q}_t^\pi$ are defined as

$$\bar{\boldsymbol{u}}_t = \frac{1}{MT}\sum_{k=1}^{T}\sum_{i=1}^{M}\mathbf{U}_{t,i}^{(k)} \quad (22)$$

$$\bar{Q}_t^\pi = \frac{1}{MT}\sum_{k=1}^{T}\sum_{i=1}^{M}Q^\pi(\hat{s}_{t,i}^{(k)}, \mathcal{A}) \quad (23)$$

where $\mathbf{U}_{t,i}^{(k)}$ and $Q^\pi(\hat{s}_{t,i}^{(k)}, \mathcal{A})$ denote the transpose of the $i^{\text{th}}$ row vector of $\mathbf{U}_t^{(k)}$ and $Q^\pi(\hat{\mathbf{S}}_t^{(k)}, \mathcal{A})$. Let $\mathbf{P}_t = \mathbf{A}_t^{-1}$. By using the Sherman-Morrison formula for (20), we can get

$$\mathbf{P}_t = \frac{1}{\lambda}(\mathbf{P}_t - \boldsymbol{g}_t\boldsymbol{v}_t^\top) \quad (24)$$

where

$$\boldsymbol{v}_t = \mathbf{P}_{t-1}\bar{\boldsymbol{u}}_t \quad (25)$$

$$g_t = \frac{v_t}{\lambda + v_t^\mathrm{T} \bar{u}_t} \quad (26)$$

Plugging (21) and (24) into (15), we finally get

$$\Theta_t \approx \Theta_{t-1} + \frac{P_{t-1}\bar{u}_t(\bar{Q}_t^\pi - \bar{Q}_t)^\mathrm{T}}{\lambda + v_t^\mathrm{T}\bar{u}_t} \quad (27)$$

where

$$\bar{Q}_t = \frac{1}{MT}\sum_{k=1}^{T}\sum_{i=1}^{M} Q(\hat{s}_{t,i}^{(k)}, \mathcal{A}; \Theta_{t-1}) \quad (28)$$

where $Q(\hat{s}_{t,i}^{(k)}, \mathcal{A}; \Theta_{t-1})$ is the transpose of the $i^\text{th}$ row vector of $Q(\hat{S}_t^{(k)}, \mathcal{A}; \Theta_{t-1})$.

ESNRLS-Sarsa is similar to ESNRLS-Q. The main difference is that ESNRLS-Sarsa is an on-policy algorithm. Similar to the deep Sarsa, $Q^\pi(\hat{S}_t^{(k)}, \hat{a}_t^{(k)})$ in ESNRLS-Sarsa is calculated as

$$Q^\pi(\hat{S}_t^{(k)}, \hat{a}_t^{(k)}) = \hat{r}_t^{(k)} + \gamma Q(\hat{S}_t'^{(k)}, a_{t+1}; \widetilde{\Theta}) \quad (29)$$

where $a_{t+1}$ is the action of the agent at $s_{t+1}$. Note that $Q(\hat{s}_{t,i}'^{(k)}, a_{t+1}; \widetilde{\Theta}) = 0$ if $\hat{s}_{t,i}'^{(k)}$ is an absorbing state.

### B. Over-fitting Prevention

It is well known that DNNs have the risk of over-fitting, since DNNs usually have too many parameters. The RLS algorithm is prone to over-fitting, too. Here we present an $L_1$ regularization method to prevent our algorithms from over-fitting. Based on Ekşioğlu's work [21], we add an $L_1$ regularization term into (27), that is

$$\Theta_t \approx \Theta_{t-1} + \frac{P_{t-1}\bar{u}_t(\bar{Q}_t^\pi - \bar{Q}_t)^\mathrm{T}}{\lambda + v_t^\mathrm{T}\bar{u}_t} - \kappa P_{t-1} sgn(\Theta_{t-1}) \quad (30)$$

where $\kappa$ is the regularization factor, and $sgn(\cdot)$ is the sign function.

### C. Overestimation prevention

In RL, policy control algorithms are prone to overestimation since they are based on boosting learning. DQN and the deep Sarsa use the target network to overcome this problem. Although our algorithms also use the target network, they still have the risk of overestimation, since part parameters of LiSEN are fixed. Here we employ the Mellowmax method, proposed by Kim et al. [18], to improve our algorithms. Define the Mellowmax function as

$$Q_t^{(k)}(c) = \frac{1}{\omega}\log\frac{1}{|\mathcal{A}|}\sum_{i=1}^{|\mathcal{A}|} e^{\omega(Q(\hat{s}_t^{(k)}, a_i; \widetilde{\Theta}) - c)} + c \quad (31)$$

where $\omega$ is the temperature coefficient, and $a_i$ is the $i^\text{th}$ action in $\mathcal{A}$. On this basis, (11) and (29) can be rewritten as follows

$$Q^\pi(\hat{S}_t^{(k)}, \hat{a}_t^{(k)}) = \hat{r}_t^{(k)} + \gamma Q_t^{(k)}(\max_{a \in \mathcal{A}} Q(\hat{S}_t^{(k)}, a; \widetilde{\Theta})) \quad (32)$$

$$Q^\pi(\hat{S}_t^{(k)}, \hat{a}_t^{(k)}) = \hat{r}_t^{(k)} + \gamma Q_t^{(k)}(Q(\hat{S}_t^{(k)}, a_{t+1}; \widetilde{\Theta})) \quad (33)$$

## IV. SIMULATION

### A. Simulation Problems and Settings

In this simulation, in order to demonstrate the effectiveness of our algorithms, we will select DQN and the deep Sarsa for comparison. They use feed-forward neural networks (FNNs) with the Adam optimization, called as FNNAdam-Q and FNNAdam-Sarsa, respectively. Considering that LiESN and RLS can preserve more information, we will validate four algorithms on MDP and POMDP tasks. The MDP task is the CartPole-v0 problem in OpenAI Gym, and the POMDP task is a modified CartPole-v0 problem without the speed information of each state. To distinguish them, we call them as MDP-CartPole and POMDP-CartPole, respectively.

The settings of networks and parameters in ESNRLS-Q and ESNRLS-Sarsa are as follows: 1) All LiESNs have three layers. Their input layers have 4 nodes for MDP-CartPole, and have 3 nodes for POMDP-CartPole. Their reservoir layers have 256 nodes with the ReLU activation function. Their output layers have 2 nodes without any activation functions. $\mathbf{W}^i$ and $\mathbf{b}^r$ are determined randomly. $\mathbf{W}^r$ is determined by the spectral radius method with the scaling factor 0.95 and the sparsity 0.25. $\mathbf{W}^o$ and $\mathbf{b}^o$ are initialized randomly. $\mathbf{H}_t^{(0)}$ and $\eta$ are fixed to $\mathbf{0}$ and 0, respectively. 2) The series length of each sample in $\mathcal{D}$ is 5. 3) $\mathbf{P}_0$ is $0.4\mathbf{I}$, $\lambda$ is 0.99999, and $\kappa$ is $10^{-5}$. 4) $\omega$ is 1.

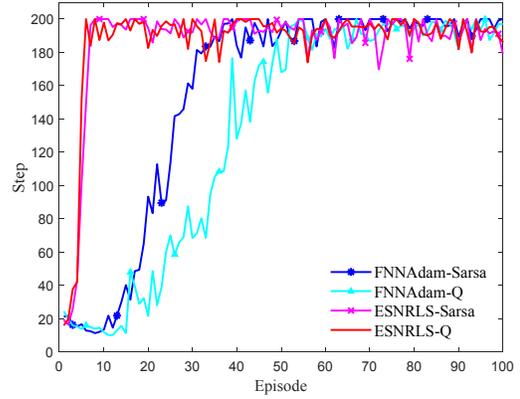

Fig.1. Performance comparison on MDP-CartPole.

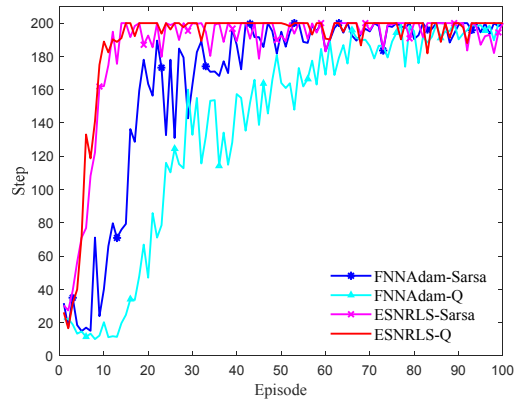

Fig.2. Performance comparison on POMDP-CartPole.

The settings of networks and parameters in FNNAdam-Q and FNNAdam-Sarsa are as follows: 1) All FNNs also have three layers. Their nodes and activation functions are the same as in LiESNs. Their weight parameters are initialized randomly, and their bias parameters are initialized to $\mathbf{0}$. 2) The hyper-parameters of Adam use the default settings in Pytorch.

The other settings of four algorithms are as follows: 1) The capacity of $\mathcal{D}$ is 100000, and the mini-batch size is 64. 2) At the beginning of each experiment, the agent runs 1000 episodes to collect enough samples with the completely random policy. 3) After collecting samples, the agent runs 100 episodes with the $\epsilon$-greedy policy. 4) $\epsilon$ and $\gamma$ are 0.01 and 0.99, respectively. 5) In each episode, the agent runs 200 time steps at most. If the running steps is less than 200, the reward of the last step is -10. 6) The target networks are updated at the end of each episode. 7) All experiments repeat 5 times.

*B. Simulation Results and Analysis*

Fig. 1 and Fig. 2 show the averaged running steps of the agent with four algorithms on two tasks. It can be seen from both figures: 1) The convergence performances of our two algorithms are obviously superior to those of FNNAdam-Q and FNNAdam-Sarsa. 2) There is almost no difference between the convergence performances of our two algorithms on the same task. Whereas, there is obvious difference between the convergence performances of FNNAdam-Sarsa and FNNAdam-Q. 3) The convergence performances of our two algorithms on the POMDP-CartPole task are slightly inferior to on the MDP-CartPole task, but they are still better than those of FNNAdam-Q and FNNAdam-Sarsa on the MDP-CartPole task. That means our two algorithms are also robust for POMDP tasks.

## V. Conclusion

Limited by the strong correlation among sequential samples of the agent, traditional ESN-based policy control algorithms use the single-series learning mode and semi-gradient descent optimization. As a result, their training efficiency and learning performances need to be improved. In this paper, we propose two novel algorithms, that is, ESNRLS-Q and ESNRLS-Sarsa. In our algorithms, we use the popular mini-batch learning mode and the LiESN model to reduce the correlation of training samples. On this basis, we present a new mean-approximation RLS method for updating LiESN to improve the training efficiency. Besides, we employ the $L_1$ regularization and the Mellowmax method to prevent our algorithms from over-fitting and overestimation. The simulation results of MDP-CartPole and POMDP-CartPole tasks show that our algorithms have better convergence performance than FNNAdam-Q and FNNAdam-Sarsa.

## Acknowledgment

This work is supported by the National Natural Science Foundation of China (61762032, 11961018). The authors are grateful for the anonymous reviewers who made constructive comments.

## References

[1] R. S. Sutton and A. G. Barto, Reinforcement Learning: An Introduction, 2nd ed. Cambridge, USA: MIT press, 2018.

[2] V. Mnih, K. Kavukcuoglu, D. Silver, A. Graves, I. Antonoglou, D. Wierstra and M. Riedmiller, "Playing atari with deep reinforcement learning," arXiv preprint arXiv: 1312.5602, 2013.

[3] V. Mnih, K. Kavukcuoglu, D. Silver, A. A. Rusu, J. Veness, M. G. Bellemare, A. Graves, M. Riedmiller, A. K. Fidjeland, G. Ostrovski, S. Petersen, C. Beattie, A. Sadik, I. Antonoglou, H. King, D. Kumaran, D. Wierstra, S. Legg and D. Hassabis, "Human-level control through deep reinforcement learning," Nature, vol. 518, no. 7540, pp. 529-533, Feb. 2015.

[4] H. Jaeger, Tutorial on Training Recurrent Neural Networks, Covering BPPT, RTRL, EKF and the "Echo State Network" Approach. Sankt Augustin, Germany: GMD-Forschungszentrum Informationstechnik, 2002.

[5] H. Jaeger, "Adaptive nonlinear system identification with echo state networks," in Proc. 15th Int. Conf. Neural Inf. Process. Syst. (NIPS), 2003: 609-616

[6] I. Szita, V. Gyenes, Lőrincz A. "Reinforcement learning with echo state networks," in Proc. 16th Int. Conf. Artif. Neural Netw. (ICANN), 2006, pp. 830-839.

[7] K. C. Chatzidimitrion, P. A. Mitkas, "A NEAT way for evolving echo state networks," in Proc. 19th Eur. Conf. AI (ECAI), 2010, pp. 909-914.

[8] I Engedy, G. Horváth, "Optimal control with reinforcement learning using reservoir computing and Gaussian Mixture," in Proc. 2012 IEEE Int. Instrum. Meas. Tech. Conf., 2012, pp. 1062-1066.

[9] D. Whitley, R. Tinós and F. Chicano, "Optimal neuron selection: NK echo state networks for reinforcement learning," arXiv preprint arXiv: 1505.01887, 2015.

[10] G. Shi, D. Liu and Q. Wei, "Echo state network-based Q-learning method for optimal battery control of offices combined with renewable energy," IET Control Theory & Applications, vol. 11, no. 7, pp. 915-922, Apr. 2017.

[11] H. Chang and K. Futagami, "Reinforcement learning with convolutional reservoir computing," Appl. Intell., vol. 50, pp. 2400-2410, Mar. 2020.

[12] M. G. Lagoudakis, R. Parr, M. L. Littman, "Least-squares methods in reinforcement learning for control," in Proc. 2nd Hellenic Conf. AI, 2002, pp. 249-260.

[13] M. G. Lagoudakis and R. Parr, "Least-squares policy iteration," J. Mach. Learn. Res., vol. 4, no. 12, pp. 1107-1149, Dec. 2003.

[14] X. Xu, D. Hu and X. Lu, "Kernel-based least squares policy iteration for reinforcement learning," IEEE Trans. Neural Netw., vol. 18, no. 4, pp. 973-992, Jul. 2007.

[15] X. Zhou, Q. Liu, Q. Fu and F. Xiao, "Batch least-squares policy iteration," Computer Science (in Chinese), vol. 41, no. 9, pp. 232-238, Sep. 2014.

[16] L. Buşoniu, D. Ernst, B. D. Schutter and R. Babuška, "Online least-squares policy iteration for reinforcement learning control," in proc. ACC, 2010, pp. 486-491.

[17] H. Jaeger et al., "Optimization and applications of echo state networks with leaky-integrator neurons," Neural Netw., vol. 20, no. 3, pp. 335-352, Apr. 2007.

[18] S. Kim, K. Asadi, M. Littman and G. Konidaris, "Deepmellow: removing the need for a target network in deep Q-learning," in Proc. 28th Int. Joint Conf. AI (IJCAI), 2019, pp. 2733-2739.

[19] D. Zhao, H. Wang, K. Zhao and Y. Zhu, "Deep reinforcement learning with experience replay based on SARSA," in Proc. 2016 IEEE Symp. Ser. Comput. Intell. (SSCI), 2016, pp. 1-6.

[20] J. Sherman and W. J. Morrison, "Adjustment of an inverse matrix corresponding to a change in one element of a given matrix," The Ann. of Math. Stat., vol. 21, no. 1, pp. 124–127, Mar. 1950.

[21] E. M. Ekşioğlu, "RLS adaptive filtering with sparsity regularization," in Proc. 10th Inf. Sci. Signal Process. Appl. (ISSPA), 2010, pp. 550-553.